\documentclass[sigconf]{acmart}
\usepackage{multirow}
\usepackage{enumitem}
\usepackage{balance}
\usepackage{fancyhdr}
\AtBeginDocument{%
  \providecommand\BibTeX{{%
    \normalfont B\kern-0.5em{\scshape i\kern-0.25em b}\kern-0.8em\TeX}}}

\setcopyright{acmlicensed}
\copyrightyear{2023}
\acmYear{2023}
\acmDOI{10.1145/3581783.3612481}

\acmConference[MM '23] {Proceedings of the 31st International Conference on Multimedia}{October 29-November 3, 2023}{Ottawa, ON, Canada}
\acmBooktitle{Proceedings of the 31st International Conference on Multimedia (MM '23), October 29-November 3, 2023, Ottawa, ON, Canada}
\acmPrice{15.00}
\acmISBN{979-8-4007-0108-5/23/10}

\begin{document}

\title{Cross-Silo Prototypical Calibration for Federated Learning with Non-IID Data}

\author{Zhuang Qi}
\orcid{1234-5678-9012}
\affiliation{%
  \institution{Shandong University}
  % \streetaddress{P.O. Box 1212}
  \city{Jinan}
  \state{Shandong}
  \country{China}
}
\email{z_qi@mail.sdu.edu.cn}

\author{Lei Meng}
\authornote{Corresponding author}
\affiliation{%
  \institution{$^1$Shandong University}
  \institution{$^2$Shandong Research Institute of Industrial Technology}
  % \streetaddress{1 Th{\o}rv{\"a}ld Circle}
  \city{Jinan}
  \state{Shandong}
  \country{China}
  }
\email{lmeng@sdu.edu.cn}

\author{Zitan Chen}
\affiliation{%
  \institution{Shandong University}
  \city{Jinan}
  \state{Shandong}
  \country{China}
}
\email{chenzt@mail.sdu.edu.cn}

\author{Han Hu}
\affiliation{%
 \institution{Beijing Institute of Technology}
 \city{Beijing}
 \country{China}
 }
\email{hhu@bit.edu.cn}

\author{Hui Lin}
\affiliation{%
  \institution{National Engineering Research Centerfor Risk Perception and Prevention}
  % \streetaddress{30 Shuangqing Rd}
  % \city{Haidian Qu}
  \state{Beijing}
  \country{China}
  }
\email{linhui@whu.edu.cn}

\author{Xiangxu Meng}
\affiliation{%
  \institution{Shandong University}
  % \streetaddress{8600 Datapoint Drive}
  \city{Jinan}
  \state{Shandong}
  \country{China}
  % \postcode{78229}
  }
\email{mxx@sdu.edu.cn}

% \affiliation{%

% \affiliation{%

\renewcommand{\shortauthors}{Zhuang Qi et al.}

\begin{abstract}
Federated Learning aims to learn a global model on the server side that generalizes to all clients in a privacy-preserving manner, by leveraging the local models from different clients. Existing solutions focus on either regularizing the objective functions among clients or improving the aggregation mechanism for the improved model generalization capability. However, their performance is typically limited by the dataset biases, such as the heterogeneous data distributions and the missing classes. To address this issue, this paper presents a cross-silo prototypical calibration method (FedCSPC), which takes additional prototype information from the clients to learn a unified feature space on the server side. Specifically, FedCSPC first employs the Data Prototypical Modeling (DPM) module to learn data patterns via clustering to aid calibration. Subsequently, the cross-silo prototypical calibration (CSPC) module develops an augmented contrastive learning method to improve the robustness of the calibration, which can effectively project cross-source features into a consistent space while maintaining clear decision boundaries. Moreover, the CSPC module's ease of implementation and plug-and-play characteristics make it even more remarkable. Experiments were conducted on four datasets in terms of performance comparison, ablation study, in-depth analysis and case study, and the results verified that FedCSPC is capable of learning the consistent features across different data sources of the same class under the guidance of calibrated model, which leads to better performance than the state-of-the-art methods. The source codes have been released at https://github.com/qizhuang-qz/FedCSPC.
\end{abstract}







\maketitle

\begin{figure}
    \includegraphics[width=0.45\textwidth]{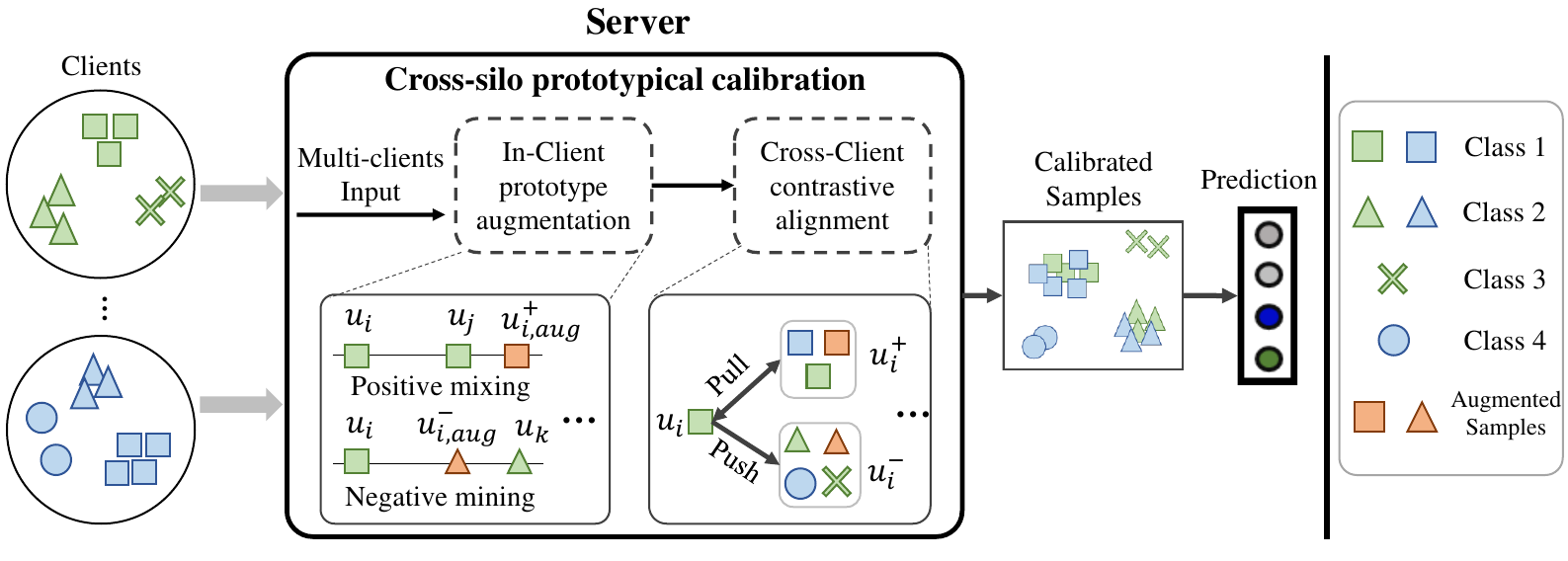}
    \vspace{-0.3cm}
  \caption{Motivation of FedCSPC. It calibrates the representation space of heterogeneous clients on the server side, which improves the generalization capability of the global model.}
  \label{fig:mg}
  \vspace{-0.3cm}
\end{figure}

\vspace{-0.1cm}
\section{Introduction}
Federated learning has gained significant attention for addressing data silos in scenarios where data sources are dispersed and difficult to share. It enables multiple parties to collaboratively train a model without sharing their data and aims to aggregate the local models obtained from the parties to generate a global model with generalization capability \cite{mcmahan2017communication,fl1,fl2}. However, the vulnerability of the federated model when confronted with heterogeneous data distribution patterns across clients has been highlighted in recent research \cite{qu2022rethinking,tang2022virtual,gao2022survey}. This is mainly due to the bias in optimization objectives among the data sources, which makes it difficult to aggregate multiple ill-posed learners into an excellent model.

To mitigate the challenge of heterogeneous data distribution, three main approaches have been developed: data sharing, mitigating the local drift on the client side  and optimizing the aggregation scheme on the server. The first method involves the use of public or synthetic datasets to create balanced data distributions, which can be beneficial in guiding clients to build unbiased models \cite{guha2019one,li2019fedmd}. The second approach typically utilizes global information as a regularizer to guide the learning process of each client, with the purpose of promoting model output consistency among clients \cite{gao2022feddc,li2020federated,li2021model,zhang2021federated,lee2022preservation,han2022fedx}. And these methods can also be divided into three subcategories: parameter-based \cite{gao2022feddc,li2020federated}, feature-based \cite{li2021model,zhang2021federated}, and prediction-based \cite{lee2022preservation,han2022fedx}. The third method considers that directly averaging parameters of local models will lead to a performance decline. They either design novel strategies to enhance the aggregation phase (such as FedMA \cite{wang2020federated}, FedNova \cite{wang2020tackling}.) or retrain the global classifier using virtual representations (CCVR \cite{luo2021no}). However, the heterogeneity of data distribution across sources results in inconsistent feature spaces, which leads to difficulties in training a model to fit data from all clients.

To address this problem, this paper presents a novel Cross-Silo Prototypical Calibration method, termed FedCSPC. As illustrated in Figure \ref{fig:mg}, compared with conventional federated learning method, the proposed FedCSPC performs prototypical calibration, which can map representations from different feature spaces to a unified space while maintaining clear decision boundaries. Specifically, FedCSPC has two main modules: the Data Prototypical Modeling (DPM) module and the Cross-Silo Prototypical Calibration (CSPC) module. To promote the alignment of features across different spaces, the DPM module employs clustering to model the data patterns and provides prototypical information to the server to assist with model calibration. Subsequently, to enhance the robustness of calibration, FedCSPC develops an augmented contrastive learning method in the CSPC module, which increases sample diversity by positive mixing and  hard negative mining, and implements contrastive learning to achieve effective alignment of cross-source features. Meanwhile, the calibrated prototypes form a knowledge base in a unified space and generate knowledge-based class predictions to reduce errors. Notably, the CSPC module is a highly adaptable tool that easily integrates into various algorithms. As observed, FedCSPC is capable of alleviating the feature gap between data sources, thus significantly improving the generalization ability.

Experiments are conducted on four datasets in terms of performance comparison, ablation study of the key components of FedCSPC, in-depth analysis and case study for the effectiveness of cross-source calibration and error analysis of FedCSPC. The results verify that FedCSPC can calibrate heterogeneous representations from different sources into a unified space via CSPC, which can mitigate the negative impact of data heterogeneity. Moreover, the error analysis reveals the potential sources of error in FedCSPC and provides insights for future improvements. 

 To summarize, this paper includes three main contributions:

\begin{itemize}[leftmargin=10pt]
    \item A novel cross-silo prototypical calibration method is proposed to alleviate the problem of data distribution heterogeneity among different clients. To the best of our knowledge, this is the first method that can map heterogeneous features from different sources to a unified space.
    \item The proposed CSPC module is an orthogonal improvement to client-based methods. Its plug-and-play design makes it easy to integrate into existing infrastructure, and it enhances the generalization without altering core components.
    \item This study reveals the fact that the inconsistent feature spaces across clients pose a challenge for the federated model to fit all clients effectively. And we have verified that FedCSPC can effectively solve this problem.
    % \vspace{-0.2cm}
\end{itemize}

\section{Related Work}

To address the issue of data heterogeneity, existing methods typically follow three main approaches: the first approach aims to alleviate the difference between local and global objectives during the local training phase, the second method focuses on optimizing the model aggregation scheme on the central server, and the third approach stems from the data sharing.

\subsection{\textbf{Models mitigating client drift}} 
\begin{figure}[t]
  \includegraphics[width=0.45\textwidth]{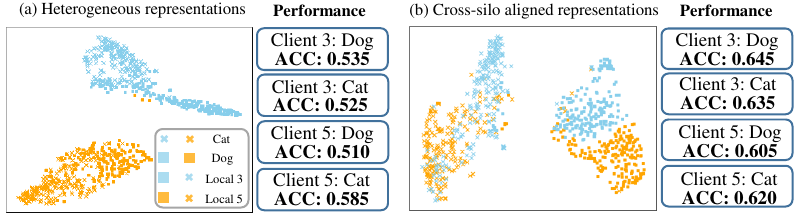}
  \caption{Feature distributions learned by FedAvg and FedCSPC. FedCSPC effectively generalizes to client-side samples by learning to calibrate client-side prototypes.}
  \label{fig:fig2}
  \vspace{-0.3cm}
\end{figure}

Common strategies in the local training phase involve utilizing global information as knowledge to regularize local updates. Conventional approaches along the line of research include weights-based \cite{gao2022feddc,li2020federated,shoham2019overcoming}, feature-based \cite{li2021model,zhang2021federated}, and prediction-based \cite{lee2022preservation,han2022fedx} constraints. Weights-based methods either design proximal terms to constrain the consistency of the local and global models or use a drift factor to track the gap between the global and local models in the parameter space. Feature-based methods focus on feature contrast to penalize inconsistency. They typically align local and global output in latent space or use prototypes to restrict clients from learning similar representations. Nevertheless, it has been observed that there exists feature maps inconsistency in these works, which leads a limited performance (See Figure \ref{fig:fig2}). Prediction-based approaches usually rely on an auxiliary dataset, and they integrate local soft-label predictions on the auxiliary dataset rather than model parameters or gradients, which reduces communication costs and achieves knowledge distillation.

 \begin{figure*}[t]
  \includegraphics[width=1.0\textwidth]{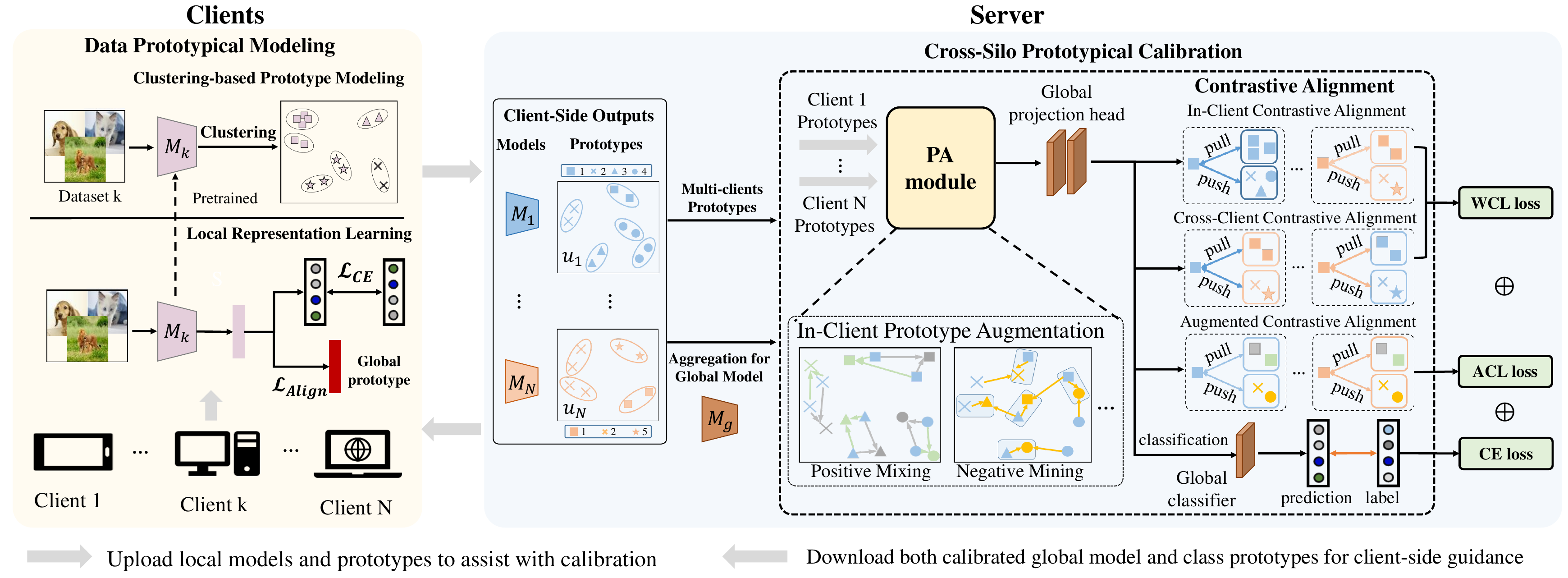}
  \vspace{-0.6cm}
  \caption{Illustration of the framework of FedCSPC. FedCSPC offers flexibility in the choice of algorithms for optimizing the local model on the client-side, including FedAvg and MOON, among others. It uses prototypes obtained from the clients to retrain the global projection head $H_g(\cdot)$ and global classifier $F_g(\cdot)$ on the server to align features from different spaces. Meanwhile, it uses prototype augmentation to improve the robustness of the calibration.}
  \label{fig:framework}
  \vspace{-0.3cm}
\end{figure*}

\subsection{\textbf{Models optimizing aggregation scheme}}
To improve the performance of the federated model, many studies focus on optimizing the aggregation mechanism on the server side. For instance, FedMA uses a Bayesian non-parametric method to match neurons rather than naively averaging \cite{wang2020federated}, FedAvgM applies the momentum rule to update the global model, which can improve robustness to heterogeneous distributed data \cite{hsu2019measuring}, and FedNova eliminates inconsistencies by normalizing local updates before averaging them \cite{wang2020tackling}. In addition, re-training or fine-tuning schemes are also applied to mitigate the model shift after aggregation, such as FedFTG uses an auxiliary generator to generate pseudo data for retraining, which can model the input space of local models \cite{zhang2022fine}. CCVR \cite{luo2021no} and CReFF \cite{shang2022federated} illustrate that the heterogeneity of the classifier is the main reason for the performance degradation of models trained on non-IID data. Therefore, they retrain the classifier by using the virtual feature generated by the gaussian mixture model and the federated feature with a consistency gradient to the real data, respectively.

\subsection{\textbf{Models Trained with Auxiliary Data}}

Due to the heterogeneity of the data distribution across different sources, the local models trained on the client side may have insufficient generalization ability for certain patterns or samples from absent classes. Therefore, existing studies propose ideas for sharing data. They typically share public datasets \cite{li2019fedmd}, synthesized datasets \cite{hao2021towards}, and truncated versions of private data \cite{guha2019one}. However, these approaches may violate privacy preservation rules since they expose the raw data to other parties.

\section{PROBLEM FORMUlation}
In federated learning, there are $N$ clients $\mathcal{C}=\{C_1, C_2, ..., C_N\}$ and a sever $S$. The client $C_k$ holds a local dataset $D_k=\{(\mathcal{X}_k, \mathcal{Y}_k)\}$ and a local model $M_k  = E_k \odot H_k \odot F_k$ with parameters $w_k=w_k^E\oplus w_k^H\oplus w_k^F$, where $E_k$ is an image encoder with parameters $w_k^E$, $H_k$ denotes projection head  with parameters $w_k^H$ and $F_k$ is a classifier with parameters $w_k^F$. The goal of federated learning methods is to jointly train a global model with the assistance of a server $S$ without leaking privacy and minimize the following problem:
\begin{equation}
    \min L(w) = \min \sum\limits_{C_k \in \mathcal{C} } {p_k L_k (w;D_k )},
\end{equation}
where $L_{k}(w)=\mathbb{E}_{(x, y) \sim \mathcal{D}_{k}}\left[\ell_{k}(w ;(x, y))\right]$ is the objective loss of $C_k$, and $
p_k  = {\textstyle{{\left| {D_k } \right|} \over D}}$ is the corresponding weight, $D = \sum\nolimits_{C_j \in \mathcal{C}} {\left| {D_j } \right|}$. After local training, clients $C_k \in \mathcal{C}$ upload the local parameters $w_j$ to sever, and the
server aggregates these parameters by 
\begin{equation} \label{eq:2}
    w_{g}  = \sum\limits_{C_k \in \mathcal{C}} {p{}_kw_k}, 
\end{equation}
The process is repeated for $T$ rounds and the resulting $M_{g}$ with the parameter $w_g$ represents the final aggregated model.

In contrast, the proposed FedCSPC introduces a \textbf{Cross-Silo Prototypical Calibration} (CSPC) module on the server, which aims to relearn the global projection head $H_g \mapsto \hat{H}_g$ the classifier $F_g \mapsto \hat{F}_g$ to align representations from different feature spaces, i.e. $
\hat H_g (E_i(x_i) ) \approx \hat H_g (E_j(x_j))$, where $x_i$ and $x_j$ are the samples with the same label in the client $C_i$ and $C_j$, respectively. FedCSPC first generates class-aware prototypes in all clients, $\mathcal{U} = \{ \mathcal{U}_k |k \in \mathcal{C} \}$ and $\mathcal{U}_k  = \{ u_k^i \left| {i \in \mathcal{Y}_k } \right.\}$ for each class on the client,  and sends them to the server. Subsequently, the CSPC module learns the mapping $\hat{H}_g(\cdot)$ based on these prototypes and the corresponding augmented samples $\mathcal{U}_{aug}$ to gather together cross-source features shared the same label. 
Finally, calibrated prototypes form a knowledge base to produce knowledge-based prediction $Pred_{k}$. The final prediction $Pred_{final}$ is achieved by $Pred_{k} \oplus  Pred_{net} \mapsto Pred_{final}$, where $Pred_{net}$ is the prediction of network.

\section{APPROACH}
\subsection{Overall framework}

FedCSPC introduces a cross-silo prototypical calibration method to enhance the generalization capability of the global model. Figure \ref{fig:framework} illustrates the main framework of FedCSPC. It first designs a novel Data Prototypical Modeling (DPM) module, which is used to model the representation distribution on the clients and provide prototypical information to the server. Afterward, the Cross-Source Prototypical Calibration (CSPC) module obtains prototypical representations from all clients and learns the mapping from the dispersed space to a unified space based on these prototypes to eliminate feature heterogeneity in the heterogeneous space. This enables the global model to generalize to all clients. Meanwhile, the calibrated prototypes form a knowledge base to aid decision-making.

\begin{figure}[t]
  \vspace{-0.3cm}
 \includegraphics[width=0.45\textwidth]{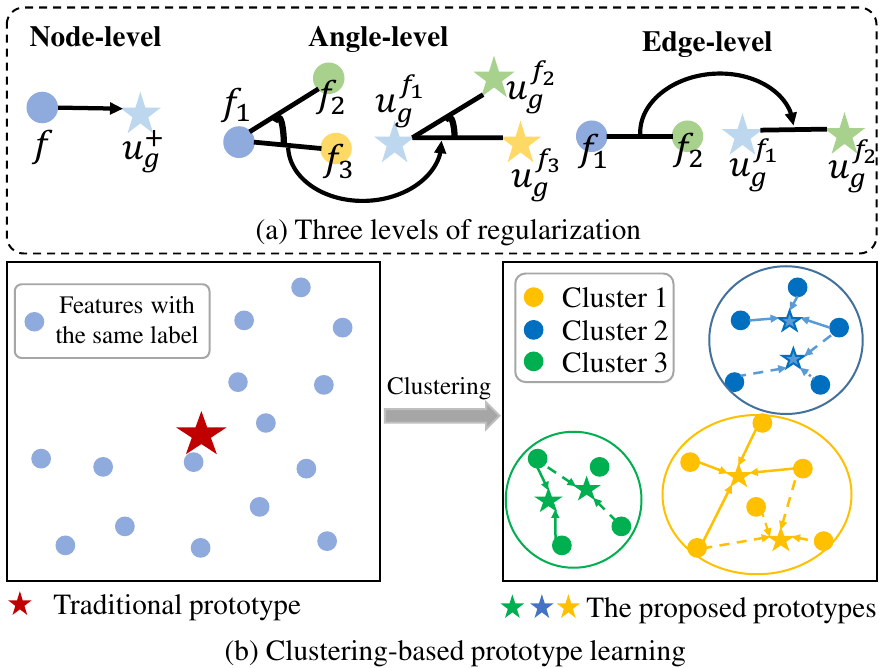}
 \vspace{-0.3cm}
  \caption{(a) Three levels of regularization to guide local representation learning. (b) Clustering-based prototype learning method better fits the distribution patterns of samples than traditional methods.}
  \label{fig:sec_4_2}
  \vspace{-0.5cm}
\end{figure}

\subsection{Data Prototypical Modeling (DPM) module}
The Data Prototypical Modeling (DPM) module aims to provide the prototypical information regarding representations to the server, which aids in model calibration. It has two main process: Strengthening local representation learning to alleviate calibration pressure and Modeling prototypical representations for data via clustering.
\subsubsection{\textbf{Local Representation Learning} (LRL)}
It is understandable that calibration becoming more difficult as the heterogeneity of features among clients increases. To alleviate this problem, the DPM module employs global prototypes $\mathcal{U}_g$ to guide all clients learn similar representations as much as possible within fewer training epochs. It uses three levels of constraints to regularize representation learning, including node, angle, and edge levels, as shown in Figure \ref{fig:sec_4_2}(a). For the point level regularization, we define an prototype-based contrastive loss to optimize the distance between the local representation and the corresponding global prototype, 
\begin{equation}
\mathcal{L}_{N}  =  - \log \frac{{\exp (f \cdot u_g^ +  /\tau_l )}}{{\exp (f \cdot u_g^ +  /\tau_l ) + \sum {\exp (f \cdot u_g^ -  /\tau_l )} }}
\end{equation}
where $f$ denotes a local representation, $u_g^+$, $u_g^-$ are the global prototypes of the same/different class as $f$, respectively. Note that the method for calculating global prototypes will be provided in the section \ref{sec:CSPC}.  $\tau_l$ is a temperature parameter. For the angle level, given three representations $f_1, f_2, f_3$ with different labels, the corresponding prototypes are $u_g^{f_1}$, $u_g^{f_2}$, $u_g^{f_3}$, and the angle-based alignment loss is defined as 
\begin{equation}
\mathcal{L}_A  =  \left\| {(\cos \angle (f1,f2,f3),\cos \angle (u_g^{f_1} ,u_g^{f_2} ,u_g^{f_3} ))}\right\|_1
\end{equation}
where $\cos \angle (f1,f2,f3) = \left\langle {\frac{{f_1  - f_2 }}{{\left\| {f_1  - f_2 } \right\|_2 }},\frac{{f_3  - f_2 }}{{\left\| {f_3  - f_2 } \right\|_2 }}} \right\rangle 
$, $\left\langle  \cdot  \right\rangle $ denotes the inner product. For the edge level, it requires the distance between the samples to be consistent with the corresponding prototypes, 
\begin{equation}
\mathcal{L}_E  = \ell (\left\| {f_1  - f_2 } \right\|_2  - \left\| {u_g^{f1}  - u_g^{f2} } \right\|_2 )
\end{equation}
where $\ell(\cdot)$ is the $L_2$ norm.

\subsubsection{\textbf{Clustering-based Prototype Modeling} (CPM)}
After representation learning, to capture the homogeneity and diversity of representations in each class, the DPM module expands the K-means clustering approach to investigate patterns in representation distributions. Specifically, the procedure of mining different patterns of class $j$ can be formulated as:
\begin{equation}
c_j^1 ,c_j^2 ,...,c_j^{k}  = \textbf{K-means}(E (x),k),x \in D_j 
\end{equation}
where $k$ is the number of clusters. $c_j^n$ denotes the $n$-th cluster of class $j$, $E(\cdot)$ is an image encoder, $D_j$ denotes data of class $j$.

To better model the distribution of representations, we repeat the process of randomly sampling $n_{repeat}$ times within each cluster to generate multiple class-aware prototypes, as shown in Figure \ref{fig:sec_4_2}(b). Compared with using a single prototype to represent the entire class distribution  \cite{mu2023fedproc,fl2}, the DPM module is capable of increasing the diversity of prototypes and providing sufficient and effective information for calibration. The calculation of the prototype can be formulated as: 
\begin{equation}
   u_j^{i,t}  = \textbf{mean}\{ f\left| f \right. \in \textbf{sampling}(c_j^i ,r)\} 
\end{equation}
where $u_j^{i,t}$ represents the $t$-th local prototype of cluster $c_j^i$, $\textbf{mean}(\cdot)$ is a Mean operation, $\textbf{sampling}(c_j^i, r)$ denotes randomly select sample features with a proportion of $r$ in cluster $c_j^i$. Finally, client $k$ sends the local model $M_k$ and local prototype set $\mathcal{U}_k=\{u_j^{i,t}|j \in \mathcal{Y}_k, i=1,2,...,n_j,t=1,...,n_{repeat}\}$ as output to the server.

\subsection{Cross-Silo Prototypical Calibration (CSPC) module} \label{sec:CSPC}
The CSPC module obtains all local models and local prototype set $\{\mathcal{M},\mathcal{U}\} =\{(M_k, \mathcal{U}_k)|k=1,...,N\}$ from clients. It has been found that regularization of client representation learning cannot completely eliminate heterogeneity. Therefore, the CSPC module is designed to align these prototypical features from heterogeneous spaces, i.e. $ \hat H_g (\mathcal{U}_i ) \approx \hat H_g (\mathcal{U}_j)$. Another challenge of learning the generalization mapping $\hat H_g(\cdot)$ to align features of heterogeneous spaces is the insufficient amount of data. Therefore, the FedCSPC develops an augmented contrastive learning method in the CSPC module. It has two main process: In-client prototype augmentation for information supplementation and Cross-client contrastive alignment for mitigating heterogeneity.

% \begin{figure}[t]
% \vspace{-0.4cm}  \includegraphics[width=0.45\textwidth]{sec_4_3.pdf}
%   \caption{Illustration of the generation of augmented samples via extrapolation and interpolation.}
%   \label{fig:sec_4_3}
%    \vspace{-0.3cm}
% \end{figure}

% , as shown in Figure \ref{fig:sec_4_3}
\subsubsection{\textbf{In-Client Prototype Augmentation} (PA)}
To augment the local prototype set, two strategies are used to generate new sample features, i.e. positive mixing and negative mining. Specifically, we use extrapolation between prototypes of the same class and interpolation between prototypes of different classes to generate positive samples and mine hard negative samples, respectively, i.e.,
\begin{equation}
u_i^{+} = (u_j  - u_i ) \times \lambda_u  + u_j , \quad u_i^{-} = (u_k  - u_i ) \times \lambda_u  + u_i 
\end{equation}
where $u_i$ and $u_j$ share the same label, whereas $u_k$ has a distinct label. $\lambda_u$ is a constant coefficient. Notably, intra-class extrapolation preserves the core features while increasing diversity. Meanwhile, inter-class interpolation injects positive information into negative samples, making it more difficult for the model to distinguish the decision boundary, which is advantageous for improving the generalization ability of the model.

\subsubsection{\textbf{Cross-Client Contrastive Alignment} (CA)}
For the raw global model $M_g=E_g  \odot  H_g \odot F_g$, obtained by Eq. \eqref{eq:2}, the projection head $H_g(\cdot)$ and $F_g(\cdot)$ need to be calibrated, i.e. $H_g(\cdot) \mapsto \hat{H}_g(\cdot)$, $F_g(\cdot) \mapsto \hat{F}_g(\cdot)$. To enhance the robustness of calibration, augmented samples are used as additional constraints,
\begin{equation}
\resizebox{0.9\hsize}{!}{$\mathcal{L}_{ACL}(u_i,u_i^{+},u_i^{-})=||H_g(u_i)-H_g(u_{i}^{+})||_2^2-||H_g(u_i)-H_g(u_{i}^{-})||_2^2+\alpha$}
\end{equation}
where $\alpha$ is the margin parameter. For the real samples, we maximize the similarity between prototypes of the same class from different sources via weighted contrastive learning, 
\begin{equation}
\resizebox{0.85\hsize}{!}{${\mathcal{L}}_{WCL} (u_i ) =  - {\textstyle{1 \over {P(u_i )}}}\sum\limits_{u_j  \in P(u_i )} {\log \frac{{\sigma _j  \cdot \exp (z_i^T  \cdot z_j /\tau _g )}}{{\sum\limits_{u_k  \in I_s } {\sigma _k  \cdot \exp (z_i^T  \cdot z_k /\tau _g )} }}} $}
 \end{equation}
where $P(u_i )$ indicates the positive set of $u_i$. $I_s$ denotes the sample set. $z_i = H_g(u_i)$, $\tau_g$ is a temperature parameter. $\sigma_j$ is a weighting factor. Considering that it is more difficult to pull samples from different sources closer and push samples from the same source farther away, we design the following rules: if $u_i$ and $u_j$ are samples of the same class from different clients or samples of different classes from the same client, $\sigma_j=1$; otherwise, $\sigma_j=0.5$. 

Meanwhile, to enhance the classification capability, the cross-entropy loss is used to further optimize the classifier $F_g(\cdot) \mapsto \hat{F}_g(\cdot)$,
\begin{equation}
\mathcal{L}_{sup}(u_i) = -\sum\nolimits_{j=1}^{N_c} \mathcal{I}(y_i=c)  \log(\hat{y}_{i,j})
\end{equation}
where $\mathcal{I}(\cdot)$ denotes the indication function, $N_c$ represent the number of classes. $y_i$ is the label of $u_i$, $\hat{y}_{i,j}$ is the prediction that $u_i$ belongs to class $j$. 

In addition, FedCSPC is unique in that it generates an exemplar $e_i$ for each class in the unified space, which serves as a knowledge base to form a knowledge-based prediction. And the final prediction of the test sample $x$ is obtained by fusing the decisions from both the network $Pred_{net}(x)$ and knowledge base $Pred_{k}(x) $, i.e.,
\begin{equation}
e^i  = \frac{1}{N}\sum\nolimits_{j = 1}^N \frac{1}{{n_{repeat} }}\sum\nolimits_{t = 1}^{n_{repeat} } {\hat H(u_j^{i,t} )}  
\end{equation}
\vspace{-0.2cm}
\begin{equation}
\resizebox{0.9\hsize}{!}{$
    Pred_{final} (x) = (1 - \lambda_p ) \times Norm(Pred_{net} (x)) + \lambda_p  \times Norm(Pred_{k} (x)) $}
\end{equation}

% \begin{equation}
% Pred_{final}(x)  = Pred_{net}(x)  \oplus Pred_{knowledge}(x) 
% \end{equation}
where $Pred_{k}(x)=[sim(f_x,e_i)\left| {i=1,...,N_c} \right.]$ contains the similarity between the sample feature $f_x$ and all exemplars $\{e_i\left| {i=1,...,N_c} \right\}$, $sim(\cdot)$ and $Norm(\cdot)$ denote the similarity and normalization function, respectively.

Furthermore, to reduce the heterogeneity of features among clients, the CSPC module generates global prototypes $\mathcal{U}_g=\{u_g^i|i=1,...,N_c\}$ to regularize the representation learning of all clients, 
\begin{equation}
u_g^i  = \frac{1}{N}\sum\nolimits_{j = 1}^N {\frac{1}{{n_{repeat} }}\sum\nolimits_{t = 1}^{n_{repeat} } {u_j^{i,t} } } 
\end{equation}
Finally, the CSPC module sends the calibrated global model $\hat{M}_g=E_g\odot \hat{H}_g \odot \hat{F}_g
$ and the global prototypes $\mathcal{U}_g$ to all clients.

\subsection{Training Strategies}
FedCSPC focuses on calibrating feature space on the server side, which can be combined with multiple client-based methods. Consequently, FedCSPC has the following training strategies.

\begin{itemize}[leftmargin=10pt]
    \item \textbf{In the client}, the optimization objective varies depending on the base algorithm being used. Moreover, the alignment loss $\mathcal{L}_{align}=\mathcal{L}_N+\mathcal{L}_A+\mathcal{L}_E$ is used to regularize all clients to learn similar representations, which can alleviate the calibration difficulty due to heterogeneity. Therefore, the overall optimization objective for a client is 
    \begin{equation}
        \mathcal{L}_{client}= \mathcal{L}_{base} +\kappa  \times \mathcal{L}_{align}
    \end{equation}
    where $\kappa$ is a weight parameter, the base algorithm could be FedAvg, FedASAM, and so on.
    \item \textbf{In the server}, FedCSPC aims to align features in heterogeneous spaces to eliminate heterogeneity and obtain clear decision boundaries, and it optimizes the following objective function:
    \begin{equation}
    \resizebox{0.90\hsize}{!}{$    \mathcal{L}_{server}  = \frac{1}{{\left| {I_s } \right|}}\sum\limits_{u_i  \in I_s } {\mathcal{L}_{\sup }(u_i)  + \eta [\mathcal{L}_{WCL}(u_i) +  \mathcal{L}_{ACL}(u_i,u_i^+,u_i^-)] }  $}
\end{equation}
where $\eta$ is a weight parameter.
\end{itemize}

\begin{table}[t]
\centering
\vspace{-0.2cm}
\caption{Statistics of CIFAR10, CIFAR100, TinyImagenet, and VireoFood172 datasets used in the experiment.}
\vspace{-0.3cm}
\begin{tabular}{c|c|c|c}
\hline
Datasets              & \textbf{\#Class} & \textbf{\#Training} & \textbf{\#Testing} \\ \hline
\textbf{CIFAR10}      & 10               & 50000               & 10000              \\ \hline
\textbf{CIFAR100}     & 100              & 50000               & 10000              \\ \hline
\textbf{TinyImagenet} & 200              & 100000              & 10000              \\ \hline
\textbf{VireoFood172} & 172              & 68175               & 25250              \\ \hline
\end{tabular} \label{tab:data}
\vspace{-0.45cm}

\end{table}

\vspace{-0.2cm}
\section{Experiments}
\subsection{Experiment Settings}
\subsubsection{Datasets}
To verify the effectiveness of the algorithms, we used four datasets in the experiment, including CIFAR10 \cite{krizhevsky2009learning}, CIFAR100 \cite{krizhevsky2009learning} and TinyImageNet \cite{le2015tiny} which are commonly used in federated learning. And a challenging food classification dataset VireoFood172 \cite{chen2016deep}. Their statistical information is shown in Table \ref{tab:data}. The Dirichlet distribution is used to partition the dataset. 

% \begin{figure}[t]
%   \includegraphics[width=0.5\textwidth]{data.pdf}
%   \caption{.... }
%   \label{fig:data}
% \end{figure}

% \usepackage{multirow}
\begin{table*}[t]
\centering
\vspace{-0.2cm}
\caption{Performance comparison between FedCSPC with baselines on CIFAR10, CIFAR100, TinyImagenet, and VireoFood172 datasets. All algorithms were run by three trials, and the mean and standard derivation are reported.}
\label{tab:pc}
\resizebox{\textwidth}{!}{
\begin{tabular}{c|c|c|c|c|c|c|c|c|c}
\hline
\multicolumn{2}{c}{\multirow{2}{*}{Methods}}                                                                      & \multicolumn{2}{|c|}{CIFAR10} & \multicolumn{2}{c|}{CIFAR100} & \multicolumn{2}{c|}{TinyImagenet} & \multicolumn{2}{c}{VireoFood172} \\ \cline{3-10} 
\multicolumn{2}{c|}{}                                                                                              & $\beta=0.1$          & $\beta=0.5$          & $\beta=0.1$           & $\beta=0.5$          & $\beta=0.1$             & $\beta=0.5$            & $\beta=0.1$             & $\beta=0.5$            \\ \hline
\multirow{8}{*}{\begin{tabular}[c]{@{}c@{}}FL without\\ Calibration\end{tabular}} & FedAvg (AISTATS'17)           & 61.18±0.7    & 66.78±0.4    & 62.12±0.8     & 66.54±0.3    & 42.48±0.8       & 45.38±0.4      & 56.78±0.9       & 59.88±0.8      \\
                                                                                  & FedProx (MLSys'20)            & 62.85±1.2    & 67.55±0.6    & 62.87±0.6     & 67.13±0.8    & 42.67±0.6       & 46.59±0.7      & 57.52±0.6       & 60.69±0.9      \\
                                                                                  & MOON (CVPR'21)                & 63.11±0.8    & 69.04±0.7    & 63.45±0.3     & 67.88±0.4    & 43.75±0.7       & 47.31±0.9      & 58.17±0.5       & 61.25±1.1      \\
                                                                                  & FedDC (CVPR'22)               & 63.25±0.7    & 69.13±0.6    & 63.76±0.7     & 67.75±0.6    & 43.68±0.8       & 46.81±0.2      & 58.04±0.4       & 60.97±0.5      \\
                                                                                  & FedNTD (NeurIPS'22)           & 62.79±0.9    & 68.89±0.3    & 62.97±1.1     & 67.83±0.2    & 43.51±1.0       & 45.79±0.5      & 57.92±0.7       & 60.88±0.9      \\
                                                                                  & FedASAM (ECCV'22)             & 63.16±0.5    & 68.48±0.6    & 63.21±0.3     & 67.71±0.5    & 43.48±0.6       & 47.38±0.6      & 58.12±0.8       & 61.14±0.2      \\
                                                                                  & Fedproc (FGCS'23)             & 62.52±1.3    & 69.18±1.2    & 63.46±1.3     & 67.63±0.7    & 43.75±0.6       & 47.21±0.4      & 57.86±0.4       & 60.46±0.4      \\
                                                                                  & FedDecorr (ICLR'23)           & 62.38±0.8    & 68.66±0.8    & 63.53±0.5     & 67.79±0.6    & 43.94±0.3       & 46.21±0.7      & 58.01±0.3       & 61.06±0.7      \\ \hline
\multirow{6}{*}{\begin{tabular}[c]{@{}c@{}}FL with\\ Calibration\end{tabular}}    & CCVR\_\{FedAvg\}(NeurIPS'21)  & 62.48±0.9    & 68.56±0.7    & 63.36±0.7     & 67.86±0.4    & 42.48±0.4       & 46.11±0.4      & 57.51±0.5       & 60.87±0.3      \\
                                                                                  & CCVR\_\{MOON\}(NeurIPS'21)    & 63.51±0.6    & 69.49±0.5    & 63.89±0.2     & 67.94±0.3    & 44.36±0.6       & 47.89±0.5      & 58.49±0.7       & 60.98±0.8      \\
                                                                                  & CCVR\_\{FedASAM\}(NeurIPS'21) & 63.12±0.8    & 69.46±0.9    & 64.18±0.6     & 68.03±0.5    & 43.73±0.3       & 47.94±0.6      & 58.79±0.5       & 61.38±0.4      \\ \cline{2-10}
                                                                                  & FedCSPC\_\{FedAvg\}           & 64.01±0.7    & 70.81±0.7    & 64.19±0.8     & 68.39±0.4    & 44.62±0.8       & 47.89±0.6      & 59.37±0.4       & 62.19±0.6      \\
                                                                                  & FedCSPC\_\{MOON\}             & 64.44±0.7    & 71.42±0.4    & 64.68±0.3     & 68.28±0.5    & 45.33±0.7       & 48.46±0.5      & 60.21±0.6       & 62.84±0.5      \\ 
                                                                                  & FedCSPC\_\{FedASAM\}          & 64.13±0.7    & 70.65±0.5    & 64.81±0.6     & 68.49±0.5    & 45.24±0.6       & 48.62±0.3      & 60.14±0.4       & 62.61±0.3    \\\hline
                                                                            
\end{tabular}}
\end{table*}

\subsubsection{Evaluation Measures} Following previous studies \cite{mcmahan2017communication,li2021model}, we use the Top-1 Accuracy to evaluate the performance of methods,
\begin{equation}
    \text { Accuracy }=(TP+TN) /(P+N)
\end{equation}
where $P$, $N$, $TP$ and $TN$ are Positives, Negatives, True Positives and True Negatives, respectively.
\subsubsection{Hyper-parameter Settings} \label{HS}
Following recent studies \cite{li2021model,mu2023fedproc}, for all methods, we set the number of clients $N=10$ with the sample fraction $C=1.0$, the number of local training epochs $E=10$, the batch size $B=64$, the communication round $T=100$ for CIFAR10 and CIFAR100 datasets, $T=50$ for TinyImagenet and VireoFood172 datasets, and the SGD optimizer with the learning rate $lr=0.01$ and the weight decay $wd$ is set to 1e-5. For all datasets, the Dirichlet parameter $\beta=0.5$ and $\beta=0.1$. In the DPM module, the number of clusters for each class $k$ is selected from $\{2,3,4\}$, the sample proportion $r=0.5$, the number of sampling $n_{repeat}=5$, and the temperature parameter $\tau_l=0.5$. In the CSPC module, the constant coefficient $\lambda_u$ and $\lambda_p$ are selected from $\{0.1,0.3,0.5\}$, the margin parameter $\alpha=1.0$, the number of augmented samples for each prototype $n_{aug}=5$, the temperature parameter $\tau_g=0.5$. For training strategies, both weight parameters $\kappa$ and $\eta$ are adjusted from $\{0.01,0.05,0.1,0.5\}$. For other compared methods, we tuned their hyper-parameters by referring to corresponding papers for fair comparison and optimal performance.

\subsection{Performance Comparison}
We compare FedCSPC with nine state-of-the-art methods, including FedAvg \cite{mcmahan2017communication}, FedProx \cite{li2020federated}, MOON \cite{li2021model}, CCVR \cite{luo2021no}, FedDC \cite{gao2022feddc}, FedNTD \cite{lee2022preservation}, FedASAM \cite{caldarola2022improving}, FedProc \cite{mu2023fedproc} and FedDecorr \cite{shi2022towards}. And the network architecture used for all methods comprises an image encoder, a projection head and a classifier. For all datasets, we employ a 2-layer MLP as the projection head and the classifier is a 1-layer fully-connected layer. For the CIFAR10 dataset, we employ a convolutional neural network comprising two 5x5 convolutional layers, which are followed by 2x2 max pooling, and two fully connected layers with ReLU function as the image encoder. For other datasets, we use a ResNet18 encoder, excluding its last fully-connected layer. The following can be observed from Table \ref{tab:pc}.
% \vspace{-0.6cm}
\begin{itemize}[leftmargin=10pt]
    \item \textbf{FedCSPC$_{\textbf{FedAvg}}$, FedCSPC$_{\textbf{MOON}}$, and FedCSPC$_{\textbf{FedASAM}}$ have demonstrated substantial improvements in classification compared to their corresponding baselines}, highlighting the model-agnostic character of the FedCSPC.
    \item \textbf{FedCSPC algorithm typically performs better than other algorithms,} which is reasonable because the calibration mechanism of FedCSPC algorithm can effectively alleviate the heterogeneity between features from different sources.
    \item \textbf{Incorporating calibration techniques into the learning process typically yields better results than the baseline method.} This is primarily because the calibration mechanism can assist devices in learning a generalized model from various data sources, such as CCVR and FedCSPC.
    % \item Generally speaking, \textbf{as the heterogeneity of data distribution increases, the performance of algorithms tends to degrade.} The proposed FedCSPC maintains its advantage over other algorithms, which may be due to its ability to model the representation distribution to assist in model calibration.
    \item As observed, \textbf{the improvement achieved by combining FedCSPC is significant compared to the baseline on CIFAR10, while it is relatively small on other datasets.} This is understandable because the final accuracy depends not only on the degree of bias correction after model calibration but also closely related to the quality of local representation learning.
\end{itemize}

\vspace{-0.2cm}
\subsection{Ablation Study}
This section further studied the effectiveness of different modules of FedCSPC. We set the sample
fraction $C=0.5$ and $C=1.0$, the Dirichlet parameter $\beta = 0.5$. The results are summarized in Table \ref{tab:as}.
% \vspace{-0.5cm}

\begin{itemize}[leftmargin=10pt]
    \item \textbf{Simply combining the traditional prototype generation method (TPG \cite{fl2}) with the cross-client contrastive alignment (CA) may not bring performance gains}, mainly because a single prototype cannot describe the overall distribution, and an insufficient number of prototypes cannot provide enough information to train a generalizable model.
    \item \textbf{Cross-client contrastive alignment (CA) with the assistance of the clustering-based prototype modeling (CPM) outperforms the base on both datasets with a large margin of up to 1.95\%, 2.11\%, 1.31\% and 0.81\%}, which verifies the effectiveness of modeling the representational distribution.
    \item In general, \textbf{using local representation learning (LRL) and prototype augmentation (PA) can further yield superior performance}, as they improve the quality of client-side representation learning and increase sample diversity, respectively, which enhances the robustness of calibration. 
    \item As reported, \textbf{knowledge-based prediction (KP) demonstrates greater efficacy on the CIFAR10 dataset compared to the CIFAR100 dataset.} This is mainly because it is easier to learn reliable classification boundaries in the representation space of CIFAR10 compared to CIFAR100.
\end{itemize}
\begin{table}[t]
\vspace{-0.2cm}
\caption{Ablation study on the effectiveness of different components of FedCSPC on the CIFAR10 and CIFAR100 datasets.}
\centering
\vspace{-0.2cm}
 \resizebox{0.48\textwidth}{!}{
\begin{tabular}{c|ccc|cc}
\hline
\multirow{2}{*}{} & \multicolumn{3}{c|}{CIFAR10}                            & \multicolumn{2}{c}{CIFAR100}      \\ \cline{2-6} 
                  & \multicolumn{1}{c|}{C=0.5} & \multicolumn{2}{c|}{C=1.0} & \multicolumn{1}{c|}{C=0.5} & C=1.0 \\ \hline
Base              & \multicolumn{1}{c|}{65.71±0.6}      & \multicolumn{2}{c|}{66.78±0.4} & \multicolumn{1}{l|}{65.01±0.7}      & 66.54±0.3 \\ \hline
+TPG+CA       & \multicolumn{1}{c|}{63.49±0.7}      & \multicolumn{2}{c|}{64.32±0.5} & \multicolumn{1}{l|}{62.37±0.5}      & 64.68±0.5 \\ 
+CPM+CA      & \multicolumn{1}{c|}{67.66±0.2}      & \multicolumn{2}{c|}{68.89±0.3} & \multicolumn{1}{l|}{66.32±0.4}      & 67.35±0.3 \\ 
+LRL+CPM+CA   & \multicolumn{1}{c|}{68.14±0.6}      & \multicolumn{2}{c|}{69.51±0.4} & \multicolumn{1}{l|}{67.04±0.1}      & 68.02±0.7 \\ 
+LRL+CPM+CA+KP   & \multicolumn{1}{c|}{68.69±0.6}      & \multicolumn{2}{c|}{69.94±0.4} & \multicolumn{1}{l|}{67.18±0.4}      & 68.14±0.3 \\ \hline
+TPG+PA+CA    & \multicolumn{1}{c|}{64.84±0.3}      & \multicolumn{2}{c|}{66.19±0.6} & \multicolumn{1}{l|}{61.79±0.6}      & 65.74±0.2  \\ 

+CPM+PA+CA   & \multicolumn{1}{c|}{68.64±0.4}      & \multicolumn{2}{c|}{69.66±0.2} & \multicolumn{1}{l|}{66.77±0.3}      & 67.77±0.4  \\ 

+LRL+CPM+PA+CA   & \multicolumn{1}{c|}{69.01±0.6}      & \multicolumn{2}{c|}{70.42±0.4} & \multicolumn{1}{l|}{67.24±0.1}      & 68.21±0.7 \\ 
+LRL+CPM+PA+CA+KP  & \multicolumn{1}{c|}{\textbf{69.47±0.3}}      & \multicolumn{2}{c|}{\textbf{70.81±0.4}} & \multicolumn{1}{l|}{\textbf{67.36±0.5}}      & \textbf{68.39±0.4} \\ \hline
\end{tabular}}
\vspace{-0.7cm}
\label{tab:as}
\end{table}

\vspace{-0.2cm}
\subsection{In-depth Analysis}

\subsubsection{\textbf{How many prototypes per cluster are enough to assist the server to calibrate a good model?}}
\begin{figure}[h]
\vspace{-0.2cm}
\includegraphics[width=0.42\textwidth]{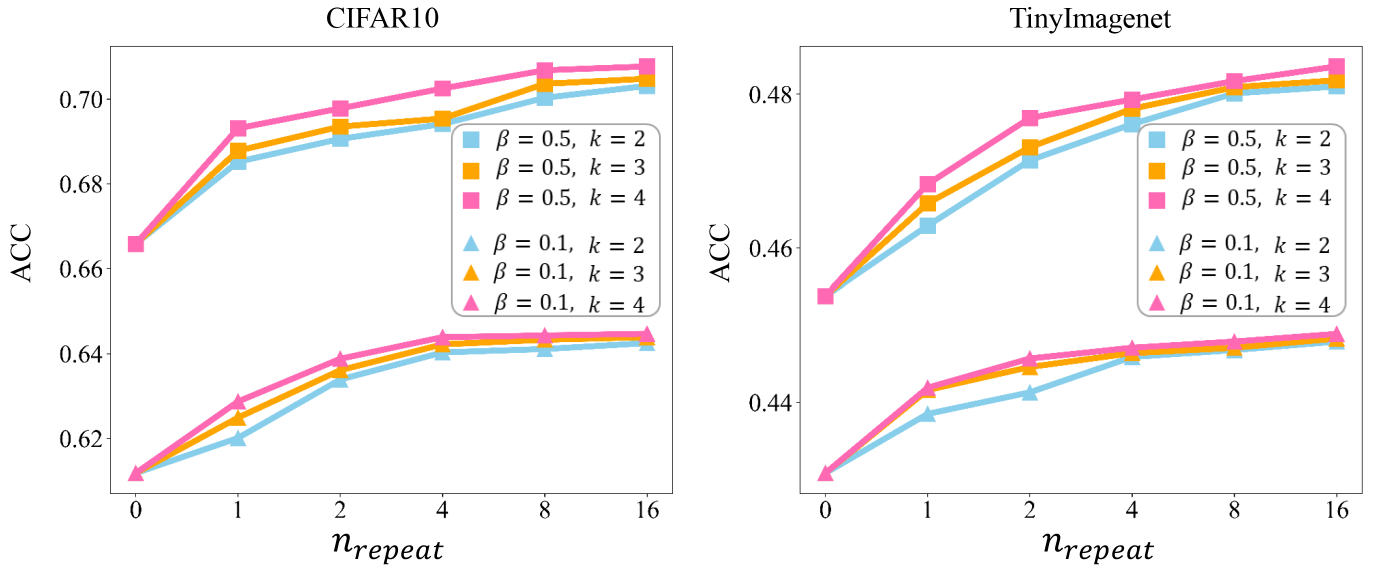}
\vspace{-0.2cm}
  \caption{The influence of the number of prototypes ($n_{repeat} = 1,2,4,8,16$) on the final performance of FedCSPC on the CIFAR10 and TinyImagenet datasets with different levels of heterogeneity ($\beta = 0.1, 0.5$) and the number of clusters $(k=2,3,4)$.}
  \label{fig:in2}
  \vspace{-0.2cm}
\end{figure}

\begin{figure}[t]
  \vspace{-0.2cm}
\includegraphics[width=0.42\textwidth]{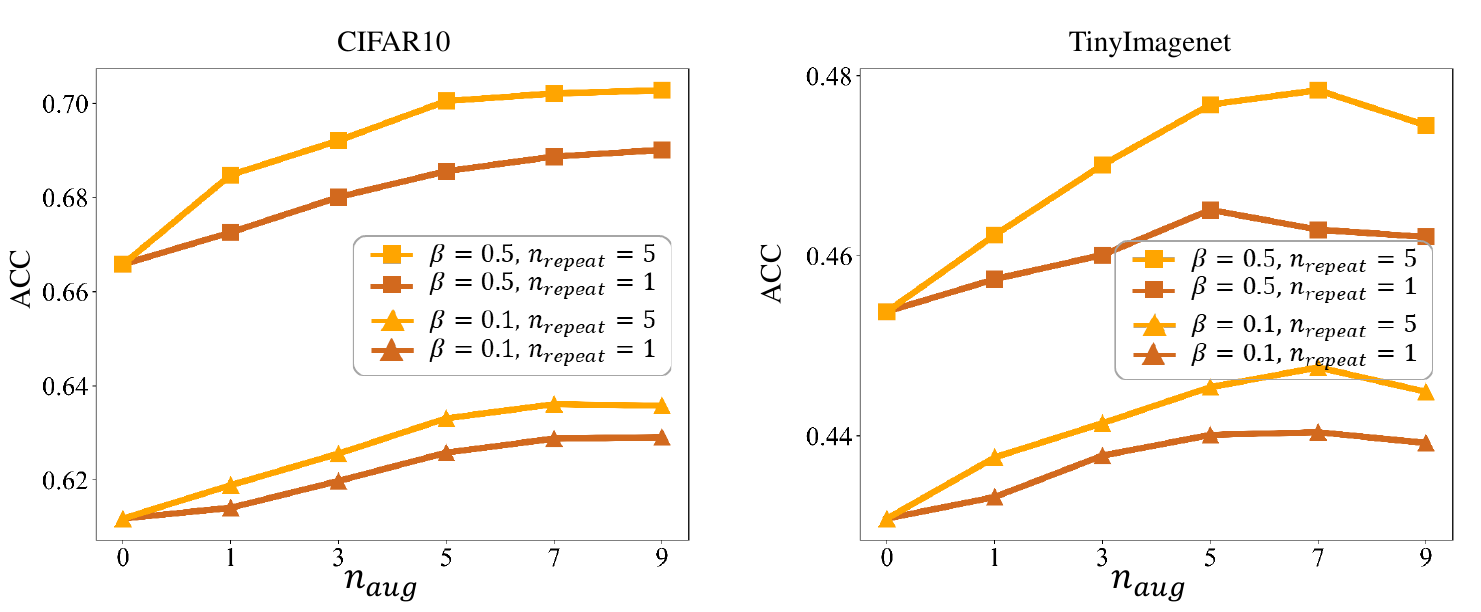}
  \vspace{-0.2cm}
  \caption{The influence of the number of augmented samples ($n_{aug} = 1,3,5,7,9$) on the final performance of FedCSPC on the CIFAR10 and TinyImagenet datasets with different levels of heterogeneity ($\beta = 0.1, 0.5$) and the number of prototypes generated per cluster $(n_{repeat}=1,5)$.}
  \label{fig:case2}
  \vspace{-0.6cm}
\end{figure}

The key hyperparameter $n_{repeat}$ is the number of prototypes generated in a cluster. We evaluate its influence by tuning it from $\{0, 1, 2, 4, 8, 16 \}$ on CIFAR10 with different heterogeneity $\beta=0.1$ and $\beta=0.5$, where $n_{repeat}=0$ denotes traditional FedAvg. And we tune the number of clusters $k$ from $\{2,3,4\}$ for both datasets. 
% Other parameters follow the section hyper-parameter settings \ref{HS}.

It can be found from Figure \ref{fig:in2} that \textbf{generating more class-aware prototypes generally leads to higher accuracy}. Moreover, with an increase in the number of prototypes, the improvement stabilizes gradually. This result is understandable, since many prototypes with high similarity are produced, which provide limited information. An impressive result is that even when only one prototype is learned per cluster, FedCSPC can still achieve an average improvement of 1.2\% and 2.2\% on CIFAR10 when $\beta = 0.1$ and $\beta=0.5$ respectively. It is worth noting that the performance of FedCSPC approach the upper limit when $n_{repeat}=4$. This finding would be conducive to mitigating the costs incurred in communication between clients and server. Additionally, \textbf{as the number of clusters generated for each class increases, the final performance gradually increases}, since clustering can effectively capture different patterns in the data, which enables the prototypes to exhibit diversity. In conclusion, although FedCSCP can bring performance gain to the baseline, the number of prototypes should be tuned carefully to achieve higher performance. 

% Typically, a larger number of prototypes is advantageous for model calibration.

\subsubsection{\textbf{How does the number of augmented samples generated for each sample affect the final performance?}}

This section explores the impact of the number of augmented samples $n_{aug}$ on the final results. We tune the $n_{aug}$ from $\{1,3,5,7,9\}$. 
We considered two cases, $n_{repeat}=1$ and $n_{repeat}=5$. 
 % Other parameters follow the section \ref{HS}.

In general, \textbf{the more augmented samples generated, the greater the performance gain for CIFAR10.} This is because the model can learn good local representations on CIFAR10, which enables the augmented positive samples to effectively increase diversity, and the augmented negative samples can help suppress overfitting, thereby enhancing the robustness of calibration. However, due to the higher complexity of the representation space in TinyImagenet, the augmented samples may contain misleading information, which increases with the number of augmented samples. This hinders the improvement of the model generalization capacity and results in a decrease in performance. Therefore, we should carefully select the number of augmented samples based on factors such as data complexity to achieve the best performance.

\begin{figure}[t]
\vspace{-0.2cm}
  \includegraphics[width=0.48\textwidth]{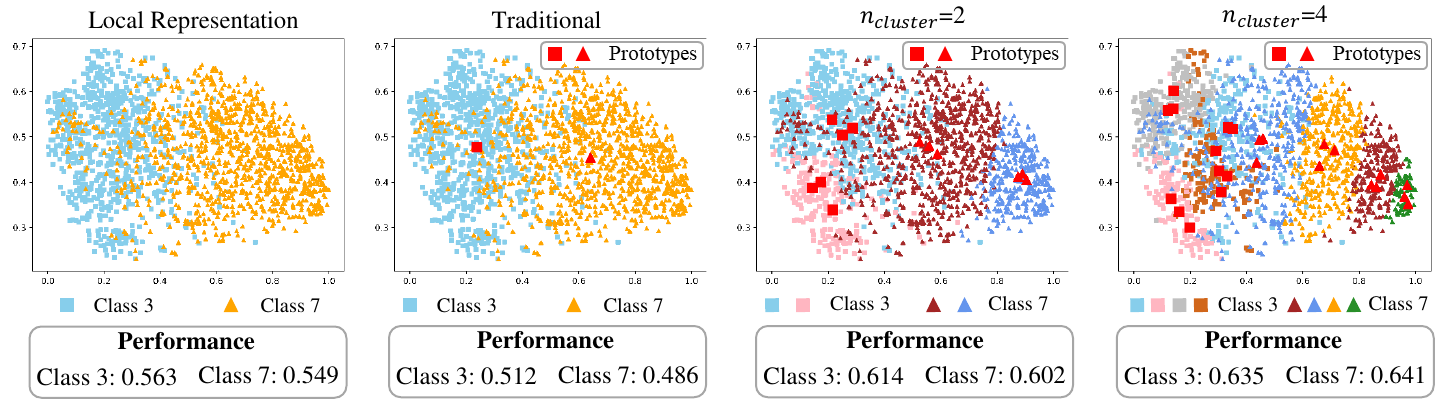}
  \vspace{-0.4cm}
  \caption{The influence of different clustering results ($n_{cluster}$=2,4 and traditional method) on the representation distribution modeling and global model performance.}
  \label{fig:case11}
  \vspace{-0.5cm}
\end{figure}

\begin{figure}[t]
  \includegraphics[width=0.45\textwidth]{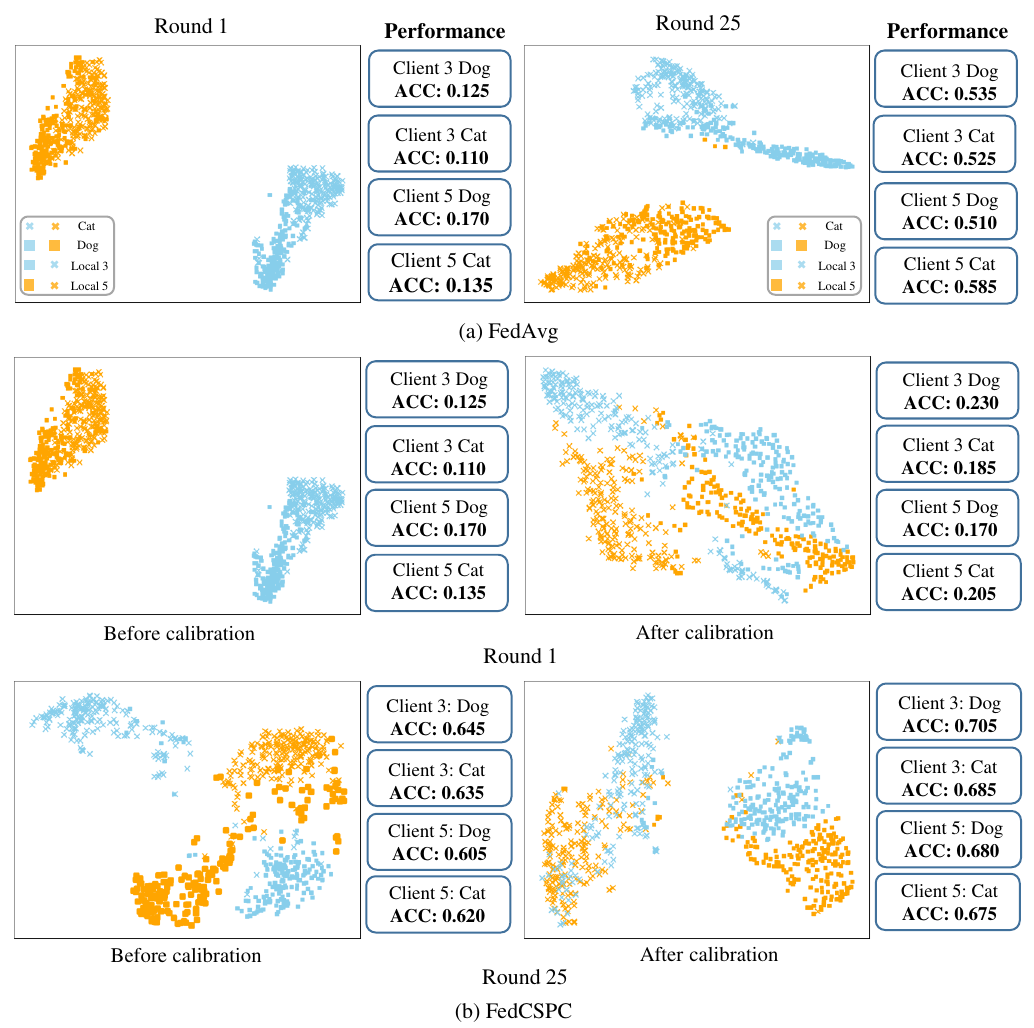}
    \vspace{-0.3cm}
  \caption{Illustration of the effectiveness of cross-silo representation alignment. (a) For the same test data, the representation distributions extracted by different local models in the FedAvg method exhibit heterogeneity. (b) FedCSPC effectively learns the common space of the same class but from different clients, which enables the global model to generalize to different clients.}
  \label{fig:case1}
  \vspace{-0.6cm}
\end{figure}

\begin{figure*}[t]
  \includegraphics[width=0.85\textwidth]{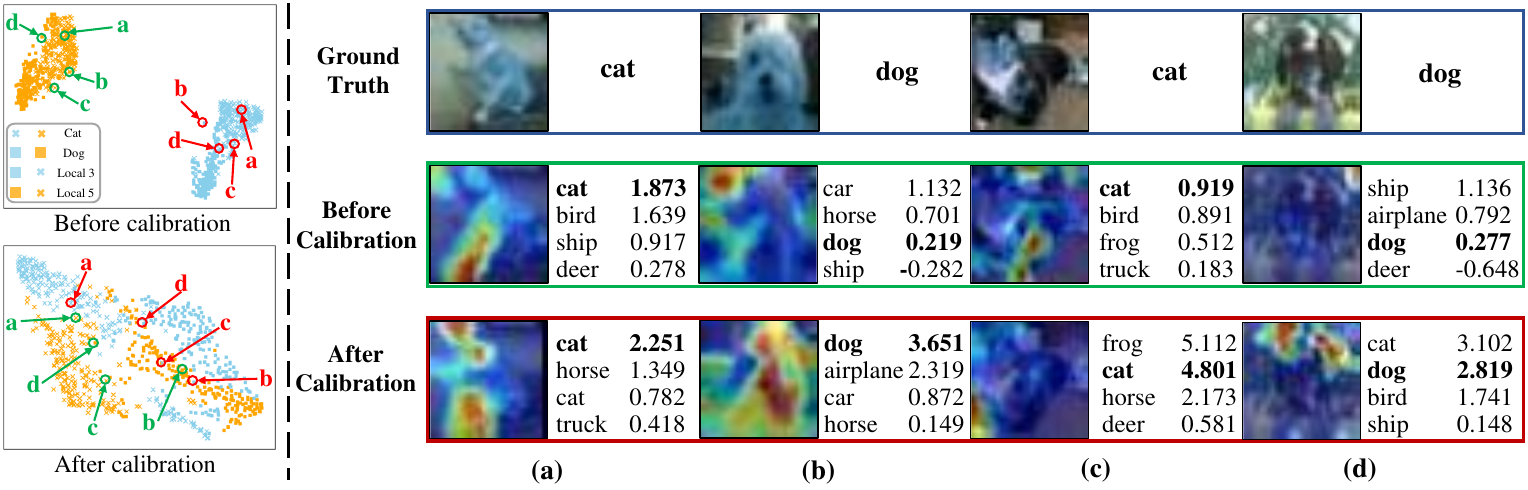}
  \vspace{-0.3cm}
  \caption{Left: the representation distribution before and after calibration. Right: the error analysis of FedCSPC. (a) FedCSPC employs cross-silo prototypical calibration method to enhance the recognition ability of the correct class  (b) FedCSPC can correct the error of prediction and calibrate the feature attention. (c) FedCSPC failed due to poor representation learning. (d) FedCSPC reduces the prediction difference between the ground-truth and top-1.}
  \label{fig:case2}
  \vspace{-0.4cm}
\end{figure*}

\vspace{-0.2cm}
\subsection{Case Study}
\subsubsection{\textbf{Clustering-based Prototype Generation}}
This section evaluates the influence of prototype generation on the representation distribution modeling and global model performance. We randomly selected the results of a training round and used TSNE \cite{van2008visualizing} to visualize the feature of two classes and their corresponding prototypes for a random user. As shown in Figure \ref{fig:case11}, \textbf{the traditional prototype cannot exhibit the intra-class diversity, while the clustering-based prototype generation method can capture the distribution patterns of representations well.} Meanwhile, we observed that clustering-based prototype generation can capture the overlapping representations between different classes (cornflower blue in the right figure). This significantly increases the diversity within each class and, as hard samples, can improve the robustness of calibration. In addition, \textbf{the more clusters generated, the more accurate the modeling of representation distribution will be, which brings more performance gains to the model ($n_{cluster}=4$ in the figure).} This is mainly because a larger number of clusters can better capture the details and diversity of sample distribution.
% However, this comes at the cost of increased complexity, which requires balancing model performance and cost.

\vspace{-0.1cm}
\subsubsection{\textbf{Cross-Silo Prototypical Calibration}} \label{QECSPC}
In this section, we randomly selected two local models and two easily confused classes (cat and dog) and extracted 200 samples from the test set for each class. The TSNE \cite{van2008visualizing} method was used to visualize the feature distribution of samples before and after calibration. We also output the corresponding classification accuracy of the model before and after calibration. As shown in Figure \ref{fig:case1}, \textbf{FedCSPC is capable of mapping features from disparate spaces to a unified space and maintaining clear decision boundaries.} In contrast, FedAvg cannot eliminate heterogeneity during training. Moreover, \textbf{FedCSPC not only corrects the distribution of heterogeneous representations in the current round but also promotes consistency in learning representations among clients.} For instance, in the 25-th round, FedCSPC almost eliminates the heterogeneity boundary of the feature distribution before calibration. This improvement in feature alignment may be a factor in the outstanding performance of FedCSPC in federated classification tasks. In addition, we note that FedCSPC was already able to calibrate heterogeneous feature distributions in the first round, but the classification accuracy remained low. This is due to the poor representation learning of local models in the 1-th round, and the limited effective information provided to the server, resulting in unreliable decision boundaries.

\vspace{-0.1cm}
\subsubsection{\textbf{Error Analysis of FedCSPC}}
This section presents a case study based on the TSNE visualization in Section \ref{QECSPC} that delves deeper into the workings of FedCSPC. To this end, GradCAM \cite{selvaraju2017grad} is employed to generate heatmaps. As depicted in Figure \ref{fig:case2}(a), both methods achieve accurate predictions for image classes. Meanwhile, FedCSPC employs calibration to attain a more precise focus on image subjects after calibration. When the target object is highly confused with other classes, the model before calibration may fail to capture the main object and make incorrect predictions. FedCSPC relies on the calibration strategy to align cross-client features (see red and green b in the left figure), which corrects prediction errors and calibrates feature attention, as illustrated in Figure \ref{fig:case2}(b). Figure \ref{fig:case2}(c) exemplifies a scenario where the model produced accurate classifications before calibration, but suboptimal representation learning hindered subsequent calibration performance, leading to an inaccurate prediction by FedCSPC (see red c in the left figure). Finally, Figure \ref{fig:case2}(d) shows the case where the model makes incorrect predictions both before and after calibration. Nonetheless, the calibration method improves the feature distribution (see red d in the left figure), which makes the model pay more attention to the dog region, reducing the discrepancy between the "dog" and the top-1 prediction. These observations not only demonstrate the effectiveness of calibration mechanisms in federated classification but also emphasize the importance of local representation learning.

\vspace{-0.3cm}
\section{CONCLUSION}
This paper presents a novel cross-silo prototypical calibration mechanism, termed FedCSPC, to handle the heterogeneity of feature space across clients. FedCSPC first employs the DPM module to mine the pattern of sample features and provide prototypical information for the server. Subsequently, the CSPC module aligns the features in the dispersed space to the unified space, and adopts prototype augmentation to improve the robustness of the alignment. Experimental results show that FedCSPC can not only calibrate the heterogeneous representation distribution of the current round, but also promote clients to learn a consistent representation in subsequent rounds and using this scheme makes FedCSPC outperform existing methods in federated classification. 

Despite the significant performance improvements achieved by FedCSPC, there are two directions that could be further explored in future work. First, stronger representation learning techniques that better learning discriminative features in clients can significantly improve performance \cite{li2021comparative,gen3,cls5,2020Multi,qi2022novel}. Second, it would be worthwhile to extend the FedCSPC to more challenging tasks, such as multimodal learning \cite{guo2020ld,meng2020heterogeneous,cl3,cls1,cls2,cls4,cls6,yang2,yang3,yang4,yang6,yang7}, recommendation \cite{RS1,RS2,RS4,RS5} and some  generative tasks \cite{wu2020multitask,gen1,gen2,gen3,gen4,gen5,gen6}.

\vspace{-0.2cm}
\section*{Acknowledgments}
This work is supported in part by the National Key R\&D Program of China (Grant no. 2021YFC3300203), the National Natural Science Foundation of China (Grant no. 62006141), the Oversea Innovation Team Project of the "20 Regulations for New Universities" funding program of Jinan (Grant no. 2021GXRC073), the Excellent Youth Scholars Program of Shandong Province (Grant no. 2022HWYQ-048), and the TaiShan Scholars Program (Grant no. tsqn202211289).

\balance
\bibliographystyle{ACM-Reference-Format}
\bibliography{sample-base}

\end{document}